\PassOptionsToPackage{table}{xcolor}
\documentclass[10pt,twocolumn,letterpaper]{article}
\usepackage{multirow}
\usepackage{float}
\usepackage{svg}
\usepackage{arydshln}
\usepackage{algpseudocode}
\usepackage{algorithm}
\usepackage[pagenumbers]{iccv} % To force page numbers, e.g. for an arXiv version

\usepackage{indentfirst}

\usepackage{pdfrender}
\usepackage{pifont}
\newcommand{\cmark}{\ding{51}}%
\newcommand{\bcmark}{\ding{52}}%
\newcommand{\xmark}{\ding{55}}%

\definecolor{tableblue}{HTML}{7FC5FF}
\definecolor{colordone}{HTML}{000000}

\definecolor{colorreview}{HTML}{000000}

\definecolor{iccvblue}{rgb}{0.21,0.49,0.74}
\usepackage[pagebackref,breaklinks,colorlinks,allcolors=iccvblue]{hyperref}

\def\paperID{5280}
\def\confName{ICCV}
\def\confYear{2025}

\title{
MIORe \& VAR-MIORe: Benchmarks to Push the Boundaries of Restoration
}

\author{George Ciubotariu
\and
Zhuyun Zhou$^{*}$
\and
Zongwei Wu
\and
Radu Timofte
\and
 \small
      Computer Vision Lab, CAIDAS \& IFI, University of W\"urzburg
\\
\small
Project Page: \textcolor{blue}{https://github.com/george200150/MIORe}
}

\begin{document}
\twocolumn[{
\renewcommand\twocolumn[1][]{#1}
\maketitle
\begin{center}
    \centering
    \includegraphics[width=\linewidth]{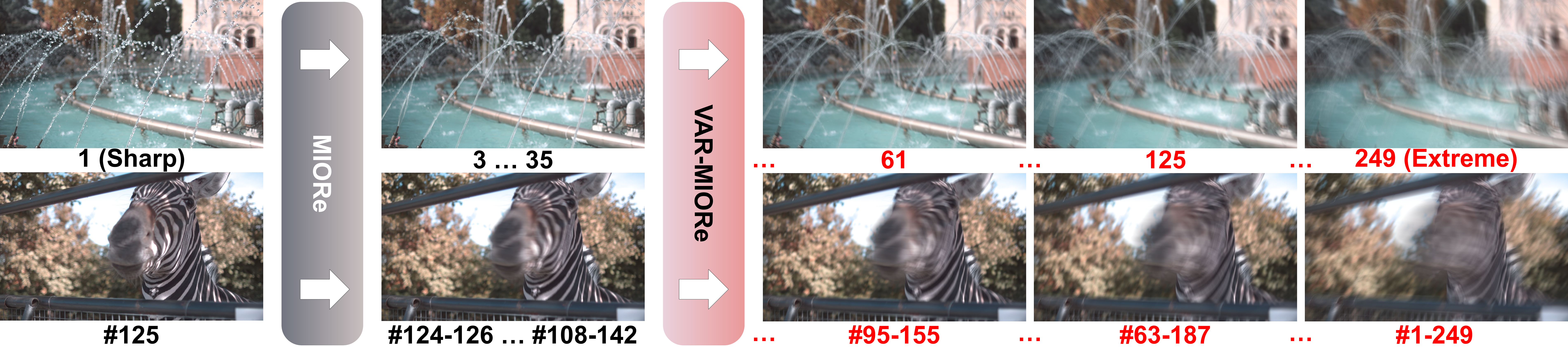}
    \captionof{figure}{
    \textbf{Continuous Motion Magnitude Range Captured by Our Novel Datasets.}
    The figure illustrates some sequences of representative frames spanning a wide dynamic range of motion magnitude (from sharp to extreme). Numerical annotations denote the integration window parameters: for instance, ``Sharp'' indicates a minimal integration span of only ``1'' frame, namely ``\#125'' yielding a sharp image, while ``Extreme'' corresponds to maximal integration, resulting in severe motion magnitude. Intermediate samples demonstrate the gradual blending of motion cues enabled by our proposed datasets. In the middle panel (``MIORe''), we aim to regularize the perceived motion magnitude inversely proportional to the scene speed, by varying the range of 3 to 35 composed sharp frames per blurry image.
    In the right panel (``VAR-MIORe''), our novel approach captures a full spectrum of motion intensities that existing works cannot achieve. These sample frames are drawn from our novel \emph{VAR-MIORe} dataset, recorded at 1000 FPS in FHD resolution, where exposure offsets from 1 to 249 ms generate progressively intense magnitude. This continuous motion variation is pivotal for advancing restoration tasks such as non-uniform motion deblurring, video frame interpolation, and optical flow estimation, in challenging real-world scenarios.
    }
    \label{fig:teaser}
\end{center}
}]

\let\thefootnote\relax\footnotetext{*Corresponding author}

\begin{abstract}

\vspace*{-5mm}
We introduce \emph{MIORe} and \emph{VAR-MIORe}, two novel multi-task datasets that address critical limitations in current motion restoration benchmarks. Designed with high‑frame‑rate (1000 FPS) acquisition and professional-grade optics, our datasets capture a broad spectrum of motion scenarios, which include complex ego‑camera movements, dynamic multi‑subject interactions, and depth-dependent blur effects.
By adaptively averaging frames based on computed optical flow metrics, \emph{MIORe} generates consistent motion blur, and preserves sharp inputs for video frame interpolation and optical flow estimation. \emph{VAR-MIORe} further extends by spanning a variable range of motion magnitudes, from minimal to extreme, establishing the first benchmark to offer explicit control over motion amplitude. We provide high-resolution, scalable ground truths that challenge existing algorithms under both controlled and adverse conditions, paving the way for next-generation research of various image and video restoration tasks.

\end{abstract}

\section{Introduction}
\label{sec:intro}

Datasets have long been the cornerstone of progress in computer vision, driving innovation by providing benchmarks that enable rigorous evaluation and comparison of algorithms. In motion restoration, which encompasses, for instance, non-uniform single-image deblurring, video frame interpolation (VFI), and optical flow (OF) estimation, the quality, diversity, and scale of available datasets directly influence the development of robust, generalizable models. Early datasets such as GoPro \cite{gopro_dset} and DVD \cite{dvd_dset} have played an important role in advancing deblurring research by providing high-quality blurred-sharp image pairs. However, these datasets are often constrained by fixed frame rates and limited motion variability, which restricts their applicability to a narrow range of motion conditions.

More recent benchmarks, for example, SINTEL \cite{sintel_dset} and FlyingChairs \cite{flyingchairs_dset} have contributed significantly to optical flow estimation by offering large-scale synthetic data with precise ground truth. However, these datasets typically do not capture the interplay between complex motion blur and dynamic scene content encountered in real-world scenarios. Similarly, while the Vimeo90K dataset \cite{vimeo90k_dset} has become a standard for video frame interpolation by curating extensive video sequences with moderate motion, its design inherently filters out extreme motion dynamics, limiting its utility to evaluate restoration under severe degradation.

The limitations of existing datasets become more pronounced when considering the increasing demand for models that can handle diverse, real-world degradations. Modern applications, ranging from autonomous driving to high-speed video analysis, require algorithms that are not only effective under controlled laboratory conditions, but also robust to a wide spectrum of motion intensities, lens-induced artifacts, and environmental variations. For instance, datasets like KITTI \cite{kitti_dset} capture real-world driving scenarios using multi-sensor platforms; but they often rely on vehicle-centric cameras with wide-angle lenses, leading to distortion artifacts that differ markedly from those seen in consumer-grade or professional cinematographic setups.

Moreover, many current benchmarks fail to address the simultaneous occurrence of motion blur and defocus blur. When a camera focuses on a salient subject, the background suffers from defocus blur, resulting in a bokeh effect that further complicates the restoration task. While several works have attempted to address defocus blur in isolation \cite{Sun_blur_kernel, Köhler_blur_kernel}, a unified framework that accounts for both motion and defocus degradations remains underexplored.

In response to these challenges, we introduce \emph{MIORe} and \emph{VAR-MIORe}, two novel multi-task datasets specifically designed to push the boundaries of motion restoration research. Our work is motivated by the need for a benchmark that captures the full complexity of real-world motion, including both controlled and adverse conditions. \emph{MIORe} is designed to represent a fair and comprehensive multi-task benchmark that supports frame interpolation, optical flow estimation, and motion deblurring. By leveraging high-frame-rate (1000 FPS) acquisition and professional-grade optics, we are able to capture sequences that exhibit complex ego-camera movements, dynamic multi-subject interactions, and depth-dependent blur effects.

A key innovation in our approach is the adaptive frame averaging mechanism. Unlike traditional datasets that generate blur by averaging a fixed number of frames, our method computes both the mean and the maximum optical flow to determine the optimal number of frames for averaging. This process yields consistent and controlled motion blur while ensuring that the left- and right-most frames, which serve as ground truth for VFI and OF estimation, remain completely sharp. The absence of blur in reference frames increases the task complexity, forcing algorithms to rely solely on subtle spatial cues for motion estimation.

\emph{VAR-MIORe} extends the capabilities of \emph{MIORe} by capturing a variable range of motion magnitudes (from minimal to extreme). This is the first dataset to explicitly incorporate the variable motion amplitude as a benchmark parameter, enabling the evaluation of algorithm performance across a continuum of motion conditions. Through careful manipulation of the frame averaging process, we create sequences that simulate both low-displacement and high-displacement scenarios, thus offering unprecedented granularity in motion analysis. Figure~\ref{fig:teaser} illustrates this variable motion magnitude, highlighting how the dataset can be used to investigate the breaking points of state-of-the-art methods.

Our acquisition is performed primarily in-the-wild, ensuring that the datasets encompass a rich variety of environmental conditions.
By utilizing an industrial-grade camera paired with an array of professional prime and zoom lenses, we introduce realistic optical degradations combined with the great variety of scene natures.
These conditions are rarely captured in consumer-grade datasets or those focused on a single application domain, thereby raising the bar for subsequent research. Furthermore, our setup enables controlled experimentation with diverse motion types, ranging from smooth panning and tilt shots to dynamic effects like dolly zoom and barrel roll, which are not simultaneously present in any existing benchmark.

The integration of these multiple dimensions, such as diverse motion scenarios, variable motion amplitude, high-resolution imaging, and flexible splitting strategies, results in a dataset that is not only large-scale but also highly versatile. Our work provides a unified platform for evaluating multiple restoration tasks within a single framework, promoting multi-task learning approaches that leverage shared information across different degradation types. Such a holistic approach is critical for developing the next generation of robust and generalizable restoration algorithms.

In summary, the contributions of our work are threefold: 
\begin{itemize}
    \item Diverse Motion Scenarios: Our datasets capture a wide range of motion dynamics, including complex ego-camera movements and interactions among multiple subjects across varying depth planes.
    \item Variable Motion Amplitude: We introduce the first benchmark to explicitly incorporate a variable range of motion magnitudes, enabling detailed analysis from minimal to extreme motion conditions. 
    \item Multi-Task Benchmarking and High Fidelity: With high-resolution imagery acquired at 1000 FPS and precise ground truth for frame interpolation, motion deblurring, and pseudo-labels for optical flow, our datasets provide a unified platform for multi-task evaluation and foster the development of advanced restoration methodologies.
\end{itemize}

By addressing critical gaps in existing benchmarks, \emph{MIORe} and \emph{VAR-MIORe} not only offer a more comprehensive evaluation environment but also pave the way for innovative research in motion analysis and restoration. Our datasets are poised to become essential resources for both academic and industrial research, stimulating advances in multi-task learning and restoration techniques that can handle the full complexity of real-world degradations.

\section{Related Work}
\label{sec:rw}

\subsection{Non-Uniform Single Image Motion Deblurring}

The GoPro dataset \cite{gopro_dset} and RealBlur \cite{realblur_dset} are seminal benchmarks in deblurring research. GoPro comprises 33 sequences where blur is synthesized by averaging 7–13 consecutive frames, despite some ghosting and compression challenges. In contrast, RealBlur employs Sony A7RM3 cameras with dual shutter speeds to generate authentic motion blur, albeit with misalignment issues due to practical constraints. Recent methods further advance the field: AdaRevD \cite{adarevd} employs a reversible multi-decoder with adaptive patch exiting to progressively disentangle degradation, achieving 34.60 dB PSNR on GoPro, while LoFormer \cite{loformer}, FFTFormer \cite{fftformer}, UFPNet \cite{ufpnet}, and NAFNet \cite{nafnet} leverage frequency-domain and adaptive architectures to improve performance. Collectively, these developments illustrate the shift toward integrating frequency-domain techniques and adaptive strategies to robustly address realistic deblurring challenges.

\subsection{Video Frame Interpolation}

Robust VFI has been driven by datasets that capture the complexities of real-world motion. The Vimeo90K benchmark \cite{vimeo90k_dset} comprises 73,171 frame triplets from 14,777 clips at 448$\times$256, curated by enforcing motion magnitude, limited intensity differences, and linear motion consistency to filter out static or highly nonlinear sequences. In contrast, the XVFI dataset \cite{x4k1000_dset} leverages Phantom Flex4K™ recordings at 4096$\times$2160 and 1,000 FPS to synthesize 240 FPS sequences, though it suffers from out-of-focus regions and limited high-velocity content. On the algorithmic side, approaches such as VFIMamba \cite{vfimamba} employ Selective State Space Models for efficient long-range modeling, while Sparse Global Matching for VFI \cite{sgmvfi} and perception-oriented methods like PerVFI \cite{pervfi} enhance intermediate flow estimation and mitigate blur artifacts. Additionally, methods incorporating unified attention mechanisms \cite{emavfi, biformer} further reduce computational overhead, underscoring ongoing innovations that leverage data diversity and algorithmic design in VFI.

\subsection{Optical Flow Estimation}

Datasets for OF estimation have long driven motion analysis research. SINTEL \cite{sintel_dset}, derived from 3D animated film, presents sequences of 50 frames exhibiting large motions, specular reflections, and various blurs, rendered at 1024$\times$436. Synthetic benchmarks like FlyingChairs \cite{flyingchairs_dset} and FlyingThings \cite{flyingthings_dset} offer large-scale data via affine transformations, while KITTI \cite{kitti_dset} captures real-world driving scenarios with multi-sensor platforms, despite challenges such as rolling shutter effects. Algorithmically, VideoFlow \cite{videoflow} integrates a tri-frame module with motion propagation for long-range consistency, FlowDiffuser \cite{flowdiffuser} iteratively refines flow via a generative denoising decoder, and MemFlow \cite{memflow} enhances real-time performance through attention-based memory aggregation. Ef-RAFT \cite{efraft} builds on RAFT with dynamic search region adjustments, and FlowFormer \cite{flowformer} employs a transformer-based approach to encode 4D cost volumes, reducing average endpoint error. These datasets and methods collectively illustrate diverse strategies to push the boundaries of OF estimation.

\section{Motivation of Our Proposed Datasets}
\label{sec:motivation_dataset}

To robustly address motion deblurring, our proposed datasets, \emph{MIORe} and \emph{VAR-MIORe}, are designed to capture the full spectrum of motion types encountered in real-world imaging. In this section, we dissect the sources and characteristics of motion into a structured hierarchy, thereby motivating the design of our datasets.

\subsection{Ego-Camera Motion}

\begin{figure}[t]

    \centering
    \includegraphics[width=\linewidth]{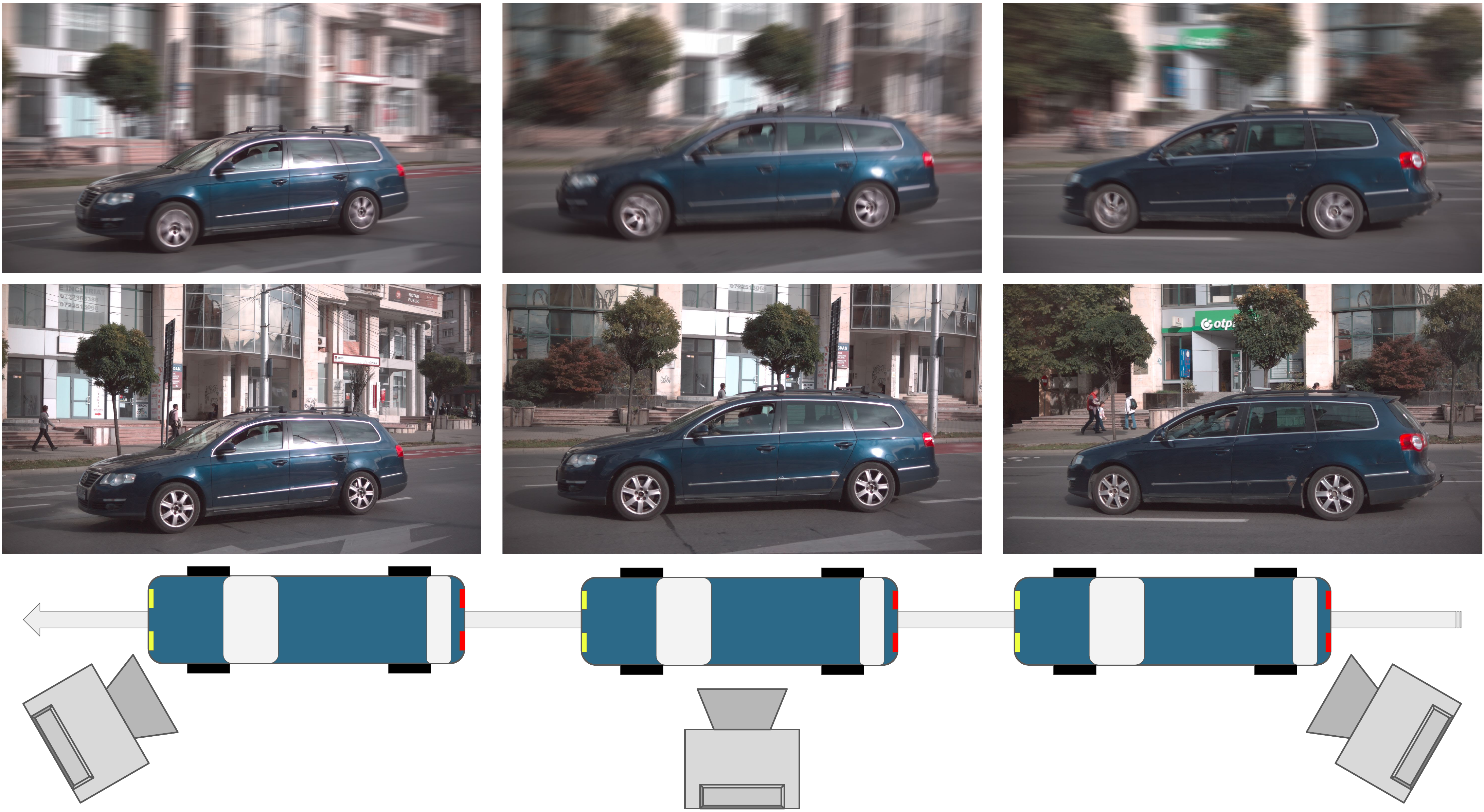}
    \captionof{figure}{
    Visualization of a panning shot from our proposed dataset \emph{MIORe} where the camera continuously tracks a moving car. The top row displays the resulting motion blur generated by the panning technique, effectively encoding the temporal evolution of the scene in a single image by integrating 19 consecutive frames. In contrast, the middle row reveals high-fidelity frames captured at 1000 FPS, which preserve fine high-frequency details despite rapid motion. The bottom row schematically delineates the dynamic camera positions relative to the car throughout the sequence, offering insight into the spatial trajectory that underpins the observed blur. This visualization underscores the complementary role of high-speed acquisition in deconstructing and ultimately restoring details from blurred imagery.
    More complex motion types are described in Figure \ref{fig:main}.
    }
    \label{fig:main2}
\vspace{-3mm}
\end{figure}

Ego-camera motion plays a crucial role in the generation of motion blur. At its simplest, \textbf{static} capture involves no movement throughout the sequence, serving as a baseline. More complex are the \textbf{Oxyz Translations} observed when the camera is mounted on a moving platform (e.g., walking or driving). In these scenarios, lateral movements produce a linear, parallel pixel shift of stationary elements, whereas \emph{zooming}, whether through lens adjustments or physical repositioning relative to the vanishing point, induces a \emph{Radial Translation} as central pixels diverge toward the edges. When both zoom-in (via lens) and zoom-out (by increasing the distance to the subject) occur simultaneously, the well-known \emph{dolly zoom} effect emerges.

Furthermore, rotational components contribute substantially to the observed blur. \textbf{YPR Rotations} are generated by either unpredictable camera shake or deliberate tracking. In the case of camera shake, rapid and chaotic rotations along the Yaw, Pitch, and Roll axes result in disordered motion blur. In contrast, smooth tracking, which is often isolated to a rotation around the yaw axis to keep a subject centered, can also involve tilt shots (pitch-only rotations) when compensating for static subjects, as illustrated in Figure \ref{fig:main2}. The roll axis is less commonly exploited, typically reserved for effects like barrel rolls in night sky cinematography.

\begin{figure}[t]
    \centering
    \includegraphics[width=\linewidth]{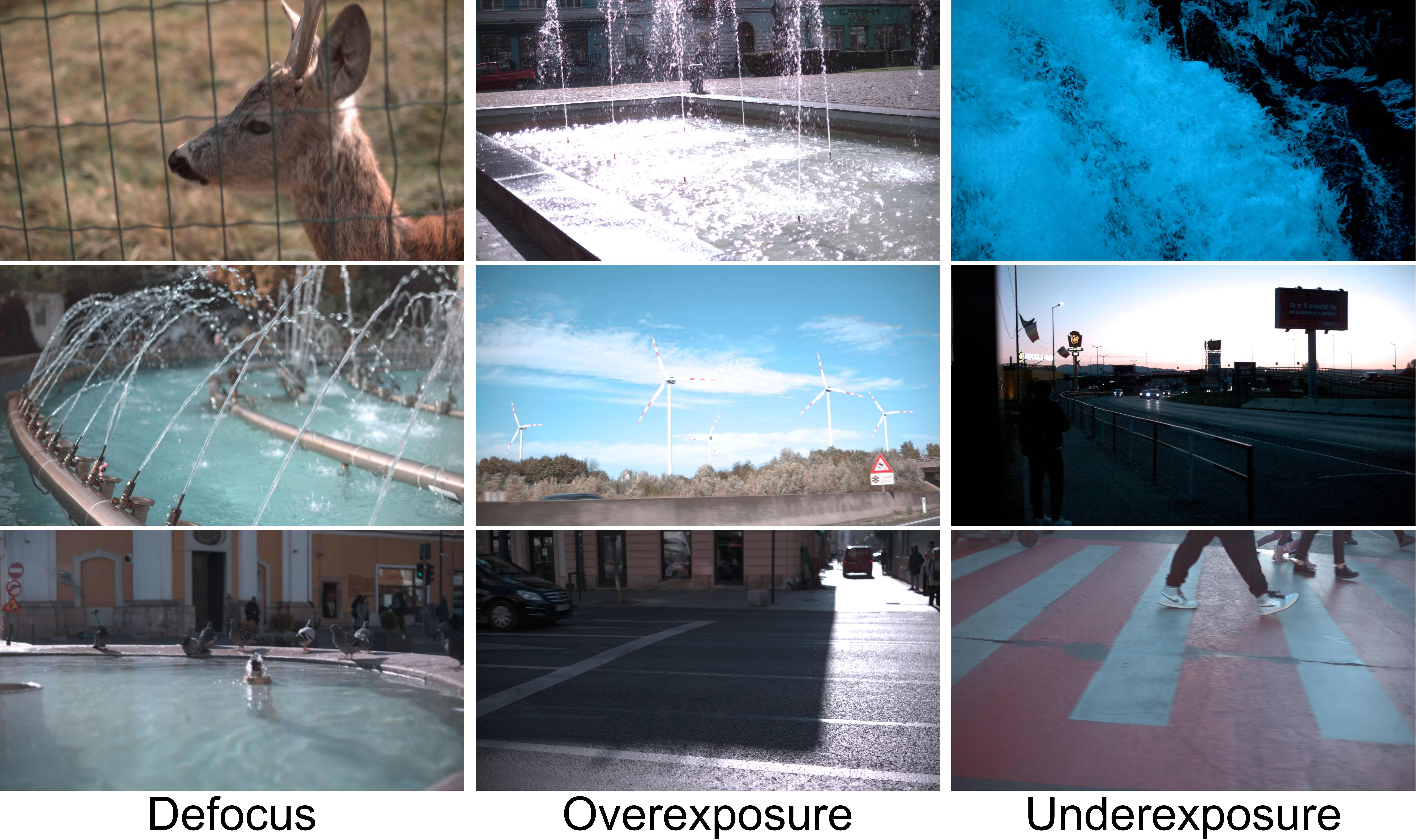}
    \vspace{-6mm}
    \captionof{figure}{
    \textbf{Adverse Imaging Conditions.} Representative examples of challenging imaging conditions in real-world captures are shown, categorized into three primary degradation types: defocus, overexposure, and underexposure. The defocus samples highlight the loss of fine details due to shallow depth-of-field (DoF), while overexposure is evidenced by blown-out whites where pixel intensities saturate, and underexposure by crushed blacks that result in diminished contrast and elevated noise levels.
    }
    \label{fig:adverse_internal}
    \vspace{-5mm}
\end{figure}

\subsection{Scene Motion}

External to ego-motion, scene motion arises from independently moving entities such as vehicles, humans, animals, and even dynamic elements like flowing liquids, fire, or turbulent smoke. Static environments also influence perceived motion. In flat backgrounds, displacements tend to be uniform except for lens-induced distortions. However, in open scenes with objects at varying depths, differential movement produces a parallax effect, which is a multi-layered phenomenon where foreground elements shift more noticeably than distant objects. This layered displacement not only complicates motion estimation, but can also lead to occlusions and depth-variable blur.

\subsection{Defocus and Extreme Exposure}
\label{ssec:defocus_exposure}

\begin{figure}[t]
    \centering
    \includegraphics[width=\linewidth]{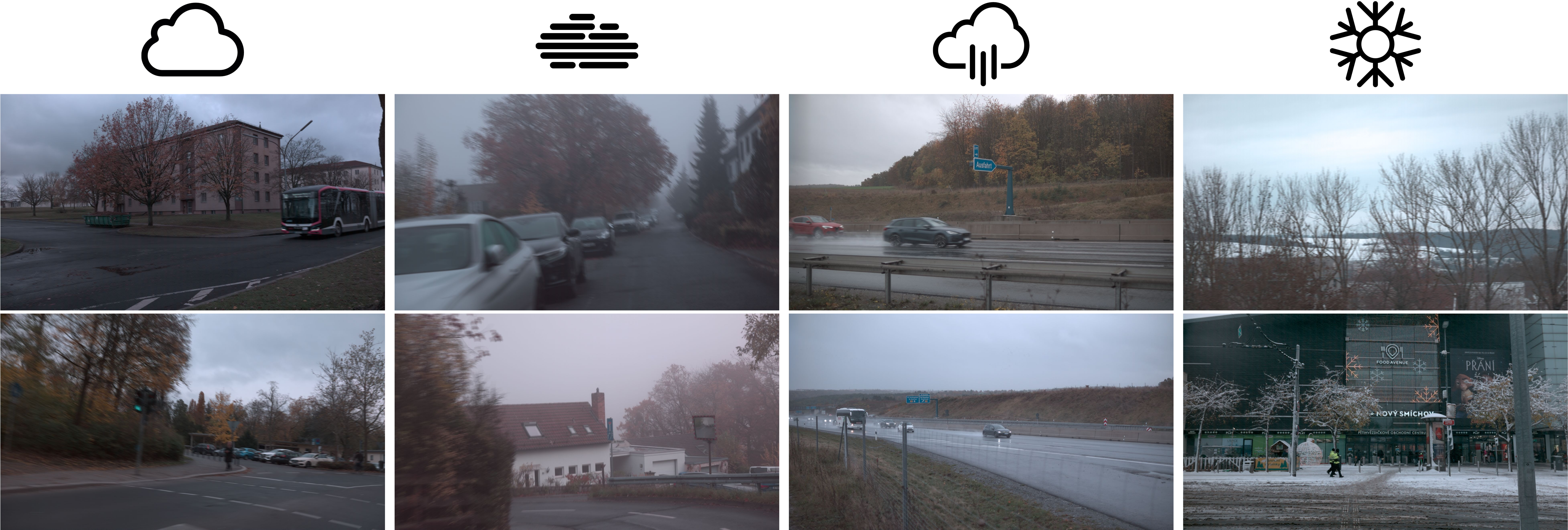}
    \vspace{-6mm}
    \captionof{figure}{
    \textbf{Variety of Adverse Conditions.} The figure displays representative examples of challenging environmental conditions encountered in real-world imaging: Clouds, Fog, Rain, and Snow. These conditions introduce distinct degradations, ranging from reduced contrast and visibility under fog and clouds to dynamic, transient artifacts in rain and snow, that pose significant challenges for tasks such as deblurring, segmentation, and restoration.
    }
    \label{fig:adverse_external}
\end{figure}

\begin{figure}[t]
    \centering
    \includegraphics[width=\linewidth]{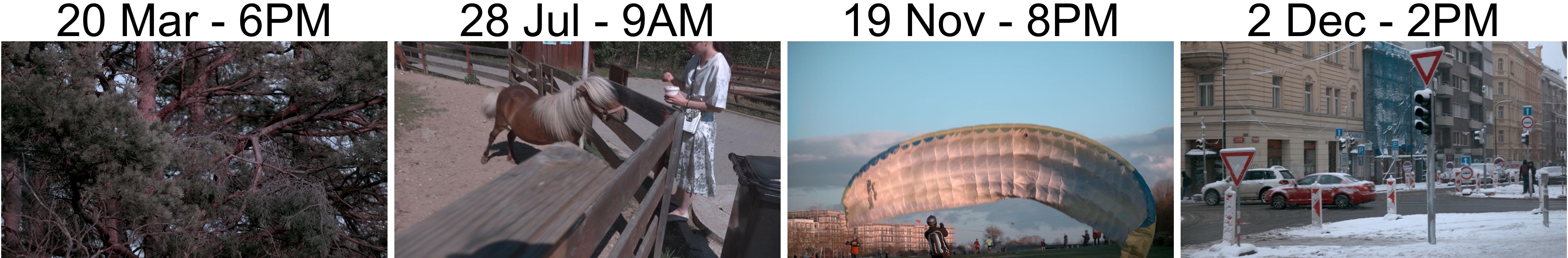}
    \vspace{-6mm}
    \captionof{figure}{
    \textbf{Time diversity: Dates and Hours}. Samples from our novel dataset, \emph{MIORe}, captured at different hours and throughout all four seasons. These images exemplify the complex lighting scenarios induced by varying sun positions, which lead to significant differences in color temperature and illumination conditions.
    }
    \label{fig:variety_date_time}
    \vspace{-3mm}
\end{figure}

When a camera focuses on salient subjects, defocus blur is inevitably introduced in the background, generating a characteristic bokeh effect that significantly alters the visual dynamics of the scene. Our proposed datasets, \emph{MIORe} and \emph{VAR-MIORe}, are the first in the motion deblurring literature to integrate defocus blur as an intrinsic component of the overall motion blur phenomenon. This comprehensive approach establishes our datasets as robust benchmarks for evaluating and developing advanced restoration methodologies capable of handling multiple, intertwined optical degradations simultaneously. As illustrated in Figure \ref{fig:adverse_internal}, these camera-specific conditions significantly increase the difficulty of image restoration by reducing the available high-fidelity information, thereby challenging conventional algorithms. Moreover, the presented diversity in degradation types provides a critical benchmark for advancing state-of-the-art restoration methods, including deblurring, exposure correction, and low-light enhancement.

\subsection{Adverse and Diverse Conditions}

Environment-specific adverse conditions also manifest as significant variations in color, white balance, and depth-dependent degradation. Figure \ref{fig:adverse_external} illustrates examples of adverse weather conditions, with each column from left to right showcasing clouds, fog, rain, and snow. This diverse presentation emphasizes the necessity for robust imaging algorithms capable of handling non-ideal acquisition scenarios, and highlights the critical need for datasets that faithfully capture the complexity of natural scenes to advance robust computer vision methodologies.

Time diversity is another crucial aspect captured in our dataset, as shown in Figure \ref{fig:variety_date_time}. Samples from our novel dataset, \emph{MIORe}, recorded at different hours and across all four seasons, reveal the complex lighting variations induced by changing sun positions. These variations lead to significant differences in color temperature and overall illumination conditions, challenging algorithms to maintain consistent performance under variable lighting. This temporal diversity not only underscores the intricacies of real-world scenes but also provides a comprehensive benchmark for developing adaptive and resilient computer vision solutions.

\section{Our MIORe \& VAR-MIORe Datasets}
\label{sec:method}

In this paper, we present two novel multi-task datasets. First, \emph{MIORe} distinguishes itself through a rich variety of motion types, ensuring that it is comparable to state-of-the-art benchmarks for non-uniform single image motion deblurring, video frame interpolation, and optical flow estimation. Second, \emph{VAR-MIORe} is designed to push the limits of movement-based tasks by covering a wide spectrum (minimal to extreme motion magnitude).
A distribution visualization of specific dataset features is presented in Figure \ref{fig:miore_metadata_stats}. More details are offered in the \emph{supplementary material}.

Moreover, we introduce a methodology for achieving regularized motion blur effects across various motion intensities by leveraging high-frame-rate (HFR) capture at 1000 FPS, as further exemplified in Figure \ref{fig:amplitude}. Our acquisition protocol enables precise ground truth (GT) generation, capturing complex motion patterns across all seasons and daylight conditions, while incorporating adverse environmental factors to enhance dataset diversity.

Motion blur is critical in visual datasets, particularly for extreme motion analysis such as high-speed tracking. Traditional datasets are often constrained by their FPS, which limits the range and quality of the OF data.
In contrast, our high-FPS raw data, processed via flexible frame averaging, results in consistent control over motion blur intensity, enabling the study of similar-magnitude blur under a broad spectrum of recorded OF conditions.

Our data is captured using an industrial-grade camera paired with a wide array of professional prime and zoom lenses. Unlike datasets shot with GoPros \cite{gopro_dset, dvd_dset, hide_dset} or vehicle-centric systems \cite{kitti2012_dset, kitti_dset}, our acquisition setup deliberately varies the lens degradation models \cite{lens_degradation, lens_degradation_noise_kernel}, vignetting \cite{lens_vignette}, and chromatic aberration \cite{lens_chr_aberr}, thereby increasing the complexity and realism of the motion-induced degradations present in our dataset.
Overall, our dataset contains 333 meticulously selected sequences, which show a great and valuable variety not only in motion granularities, but also in scenarios (natural, urban, etc.), seasons, adverse conditions; which offers a unique characteristic to our proposed dataset, and is the first to the best of our knowledge.

\begin{figure}[t]
    \centering
    \includegraphics[width=\linewidth]{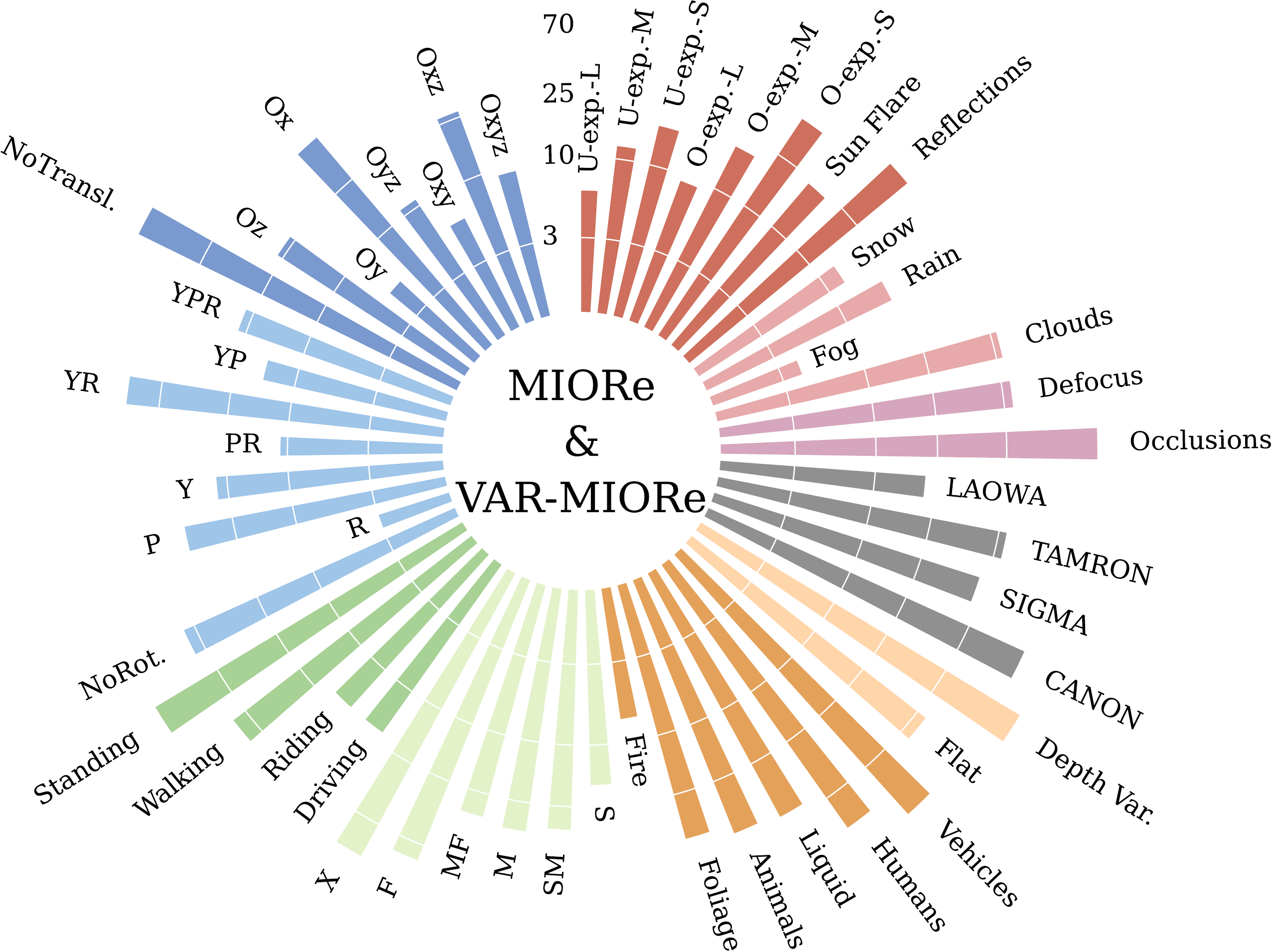}
    \caption{\textbf{Dataset sequence annotations statistics.} Our datasets contain samples exhibiting diverse motion patterns, including both subject motion and camera motion, characterized by translations along the \textbf{$Ox$}, \textbf{$Oy$}, and \textbf{$Oz$} axis, as well as rotations in \textbf{Y}aw, \textbf{P}itch, and \textbf{R}oll. These sequences also encompass a wide variety of scene content and acquisition setups.}
   \label{fig:miore_metadata_stats}
   \vspace{-3mm}
\end{figure}

\subsection{Dataset Comparison}
\label{ssec:dset_compare}

As shown in Table \ref{table:motion_types_dset_comparison}, our analysis reveals that, while popular public datasets, targeting Single Image Motion Deblurring (GoPro \cite{gopro_dset}, RealBlur \cite{realblur_dset}), Video Frame Interpolation (Vimeo90K \cite{vimeo90k_dset}, X4K1000FPS \cite{x4k1000_dset}), and Optical Flow Estimation (KITTI \cite{kitti_dset}, Sintel \cite{sintel_dset}), offer varied motion scenarios, our novel datasets, \emph{MIORe} and \emph{VAR‐MIORe}, provide comprehensive motion coverage by addressing all four motion types. Notably, \emph{VAR‐MIORe}, also recorded with most lenses (4) at highest resolution (1920$\times$1080) as our \emph{MIORe}, extends the motion magnitude from static to extreme conditions, achieving up to 1932 in Max OF and sustaining 1000 FPS for both acquisition and final take-away dataset. This extensive motion diversity, combined with minimal to maximal offsets, underpins the superior capability of our datasets for accurate motion analysis and robust restoration tasks in challenging real-world scenarios. A detailed comparison of motion types and visualization demos is provided in the \textit{supplementary material}.

\subsection{Hardware Setup and Image Acquisition}
\label{ssec:img_acq}

A CHRONOS 2.1-HD high-speed camera \cite{chronos}, capturing footage at a resolution of 1920$\times$1080 pixels (FHD) at 1000 FPS was selected as our recording device. Due to its internal RAM limitations, each continuous recording is restricted to a maximum of 5516 frames.
Additionally, for several sequences, we use a tripod for shot stabilization, limiting motion blur to only subject motion. successfully ensuring consistent framing and no motion artifacts.

\begin{figure*}[t]
    \centering
    \includegraphics[width=\linewidth]{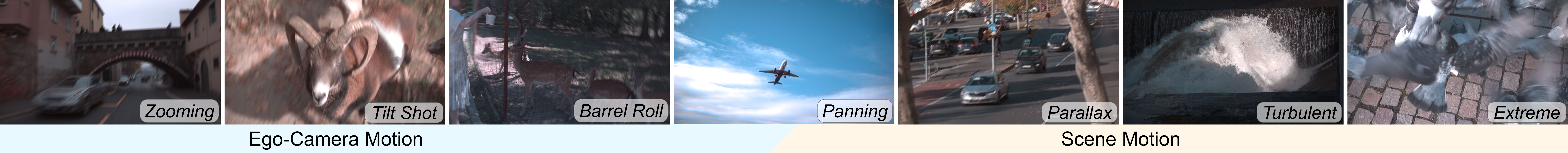}
    \vspace{-6mm}
    \captionof{figure}{
    \textbf{Diverse Motion Patterns Captured by Our Proposed Datasets.}  The figure presents a wide spectrum of motion types observed in our dataset. The top row shows samples of blurred images with ego-camera induced motions include Zooming, Tilt Shot, Barrel Roll, and Panning; and also with scene motions such as Panning and Parallax. The progression to turbulent or extreme conditions further emphasizes the dynamic range of motion captured.
    This comprehensive visualization underscores the complex interplay of various motion cues inherent to real-world scenarios. By capturing both subtle and pronounced movements, our system facilitates detailed analysis for tasks such as optical flow estimation, motion deblurring, and frame interpolation. The juxtaposition of ego-centric versus scene-centric motion also provides a critical framework for disentangling camera-induced artifacts from true scene dynamics, thereby driving advancements in robust computer vision methodologies.
    }
    \label{fig:main}
\end{figure*}

\begin{table*}[t]
\footnotesize
\begin{tabular}{l|cccc|ccccc|c}
Dataset                     & T\&R               & Tracking                 & Flt\&Prlx           & DF Obj.                  & Resolution           & \# Lens    & FPS (Acq. / Final)             & Off. (ms)           & Max OF        & Size           \\ \hline
GoPro \cite{gopro_dset}              & \cmark                        & {\color[HTML]{FF0000} \xmark} & {\color[HTML]{FF0000} \xmark} & {\color[HTML]{FF0000} \xmark} & 1280$\times$720           & 1          & 240 / {[}18, 34{]}             & {[}29, 54{]}          & 135           & 3214           \\
RealBlur \cite{realblur_dset}           & {\color[HTML]{FF0000} \xmark} & {\color[HTML]{FF0000} \xmark} & \cmark                        & {\color[HTML]{FF0000} \xmark} & 680$\times$773            & 1          & 2 / 80                         & 500                   & 47            & 4556           \\
Vimeo90K \cite{vimeo90k_dset}                    & \cmark                        & \cmark                        & \cmark                        & \cmark                        & 448$\times$256            & -          & 30 / 30                        & 33                    & 65            & 73171          \\
X4K1000FPS \cite{x4k1000_dset}                  & \cmark                        & {\color[HTML]{FF0000} \xmark} & {\color[HTML]{FF0000} \xmark} & \cmark                        & 768$\times$768            & 1          & 1000 / 240                     & 4                     & 288           & 4888           \\
KITTI \cite{kitti_dset}                       & {\color[HTML]{FF0000} \xmark} & {\color[HTML]{FF0000} \xmark} & {\color[HTML]{FF0000} \xmark} & {\color[HTML]{FF0000} \xmark} & 1242$\times$375           & 1          & 10 / 10                        & 100                   & 355           & 400            \\
Sintel \cite{sintel_dset}                      & \cmark                        & \cmark                        & {\color[HTML]{FF0000} \xmark} & {\color[HTML]{FF0000} \xmark} & 1024$\times$436           & -          & - / 24                         & 42                    & 414           & 1628           \\ \hline
\textit{\textbf{MIORe} (Ours)}     & \multicolumn{4}{c|}{- \bcmark -}                                                   &  &  & \textbf{1000 / {[}28, 1000{]}} & \textbf{{[}3, 35{]}}  & 95            & 52218          \\
\textit{\textbf{VAR-MIORe} (Ours)} & \multicolumn{4}{c|}{\textit{\textbf{- Static to Extreme -}}}                      & \multirow{-2}{*}{\textbf{1920$\times$1080}} & \multirow{-2}{*}{\textbf{4}} & \textbf{1000 / {[}4, 1000{]}}  & \textbf{{[}1, 249{]}} & \textbf{1932} & \textbf{83250}
\end{tabular}
\captionof{table}{
\textbf{Dataset Characteristics and Benchmark Metrics.}
We compare multiple datasets in terms of motion attributes and imaging performance, detailing key metrics such as Translation and Rotation (T\&R), Tracking, Flat and Parallax background motion (Flt\&Prlx), and DeFormable Object (DF Obj.) scenarios. Additional metrics include resolution, number of lenses, acquisition and final frames per second (FPS (Acq. / Final)), temporal offset between consecutive dataset samples in milliseconds (Off. (ms)), maximum optical flow (Max OF), and overall dataset size.
The comprehensive comparison between our novel datasets and other state-of-the-art (SOTA) ones underscores the superior coverage of our datasets in capturing complex and high-speed motions, which is critical for accurate motion analysis and other restoration tasks in challenging real-world scenes.
}
\label{table:motion_types_dset_comparison}
\vspace{-2mm}
\end{table*}

\textbf{Our Lens Variety}: 1) Tamron 15-30mm zoom for wide-angle to standard; 2) Canon 24mm for wide-angle; 3) Sigma 85mm for telephoto; and 4) Laowa 100mm for macro.
Each lens was used at wide-open apertures to maximize light intake, with adjustments made on bright scenes by narrowing the aperture as needed to maintain optimal exposure.

More details on the quality assurance measures implemented during the acquisition procedure are provided in the \emph{supplementary material}.

\subsection{Dataset Construction Procedure}

Our dataset construction procedure is designed to ensure high-quality ground truth for various restoration tasks. For each sequence, we determine the optimal blur intensity by computing both the mean and maximum optical flow. These metrics inform the selection of the number of frames to average when generating blurry images, while the middle frame is kept as ground truth label for the deblurring task, and the left- and right-most frames are preserved as completely sharp inputs for the other two tasks. Unlike many existing VFI/OF datasets, our approach guarantees that all input frames remain sharp, thereby increasing the task complexity as lack of any visual cues regarding the magnitude or trajectory of motion.

\begin{figure}[t]
    \centering
    \includegraphics[width=0.99\linewidth]{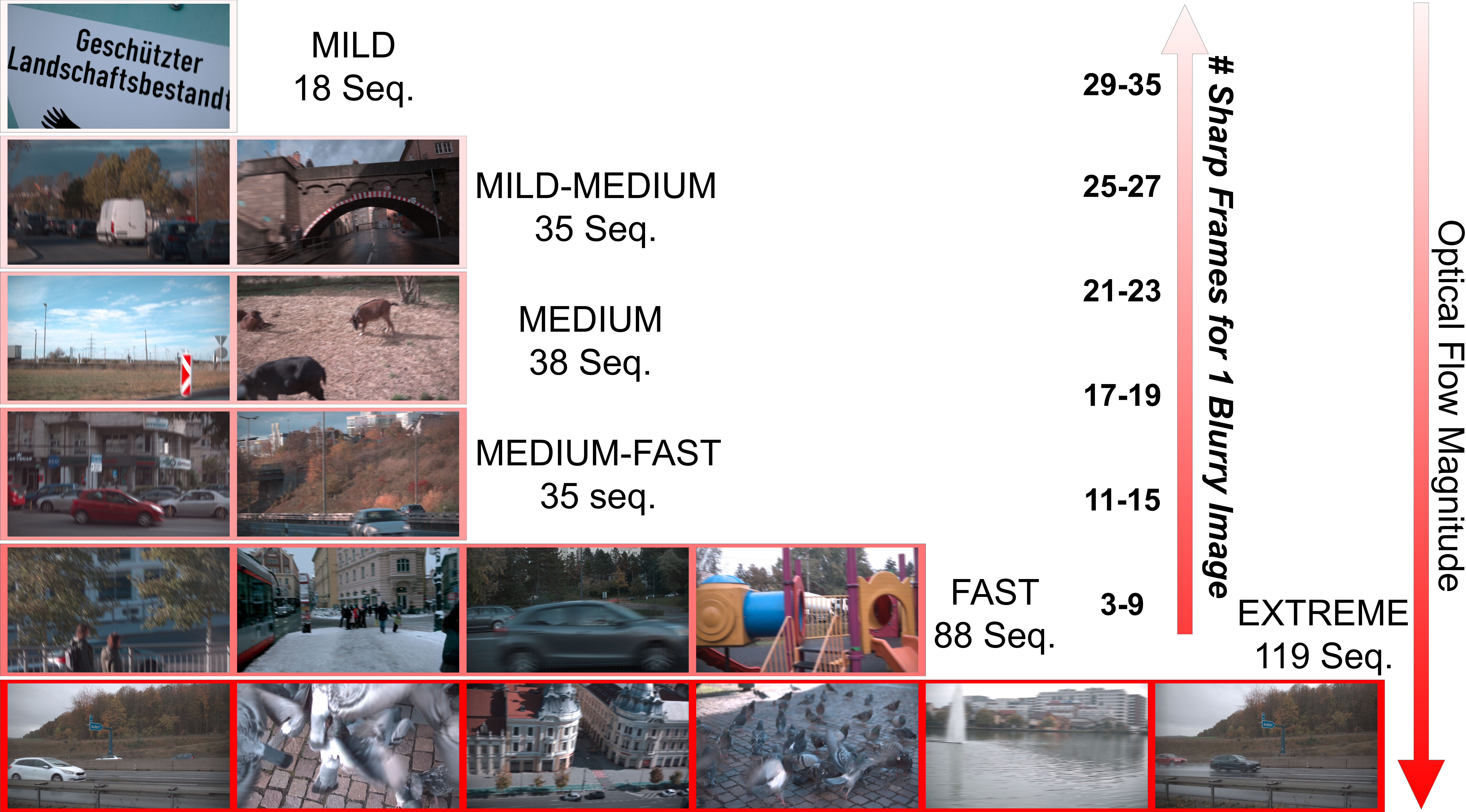}
    \captionof{figure}{
    \textbf{Motion Magnitude Distribution and Sharp-to-Blurry Frame Ratios.} We categorize sequences into six motion intensity groups, ranging from 18 MILD to 119 EXTREME and based on average optical flow magnitude, which increases progressively across categories. As motion intensifies, fewer sharp frames are averaged to synthesize each blurry frame, reflecting reduced temporal redundancy. This adaptive sampling ensures a balanced motion distribution across the MIORe-Deblurring dataset, supporting robust evaluation for deblurring, interpolation, and optical flow under realistic, high-motion conditions.
    }
    \label{fig:amplitude}
\end{figure}

We adopt established optical flow computation methodologies from prior works \cite{sintel_dset, x4k1000_dset} and enhance them with the latest approaches \cite{deqflow} to robustly capture motion characteristics. As shown in Figure \ref{fig:amplitude}, the motion magnitude or the blur intensity, is carefully proceeded: for the \emph{MIORe} dataset, we limit the range of motion to ensure comparability with existing literature, whereas \emph{VAR-MIORe} is designed to probe the ``breaking point'' of models by gradually increasing motion magnitude through extended frame averaging.

Focusing on the spatial dimension, our method disregards temporal cues and instead targets a mid-level benchmark of 30 pixels of optical flow. This is achieved by adaptively averaging frames: scenes with high optical flow (e.g., around 10 pixels per frame) are averaged over fewer frames (3 frames) to prevent excessive blur, while scenes with sub-pixel motion are averaged over a larger number of frames (up to 30) to amplify the blur effect. This adaptive strategy maintains a consistent blur level across scenes with diverse motion speeds and ensures homogeneity in the dataset.

Overall, this procedure, coupled with our use of an industrial-grade camera and a diverse set of professional lenses, produces a dataset with precise ground truth that meets the demanding requirements of advanced restoration tasks such as motion deblurring and interpolation research.

\subsection{Dataset Statistics}
\label{ssec:stats}

Both our proposed datasets are based on a total of 1,147,507 frames from the mentioned 333 selected sequences. We leverage all these frames to generate blurred frames based on the motion magnitude of the scenarios. We save the left, middle, and right sharp frames for each blurry image. For reference \emph{MIORe} consists in a total of 52,218 frames, which represent around 5\% of the original total number of sharp frames, to ensure the minimal redundancy of the data in our dataset.

For the benchmarking, we leverage the 52k blurred frames as the input for the deblurring task; for the optical flow estimation and video frame interpolation, the pairs of the starting and ending sharp frames for generating the 52k blurred frames are used as input. In the meantime, the sharp middle frame of each blurred frame is used as GT for deblurring and video frame interpolation tasks, and a carefully generated OF map serves as GT for the OF estimation task.

\section{Experiments and Benchmarking}
\label{sec:benchmark}

\subsection{Deblurring}

We benchmark our deblurring performance on both \emph{MIORe} and \emph{VAR-MIORe}. For each sequence, a single blurry image (generated by frame averaging) is used as input and compared against the corresponding sharp middle frame. Five state-of-the-art methods, all originally trained on the GoPro dataset \cite{gopro_dset}, are evaluated. Results on \emph{MIORe} are detailed in Table \ref{table:results_deblurring}, while evaluations on \emph{VAR-MIORe} are presented in Table \ref{table:psnr_varmiore_blur} and further visualized in Figures in the \emph{supplementary material}. Our analysis reveals that certain methods, such as those in \cite{adarevd, loformer, ufpnet}, deliver balanced performance across all splits, whereas others exhibit significant imbalances. Notably, \cite{fftformer} experiences a pronounced performance drop in the Mild-Medium (MS) group, and \cite{nafnet} shows poorer results in extreme motion conditions.

\begin{figure}
\centering
  \begin{subfigure}{.92\linewidth}
  \centering
    \includegraphics[width=.445\linewidth]{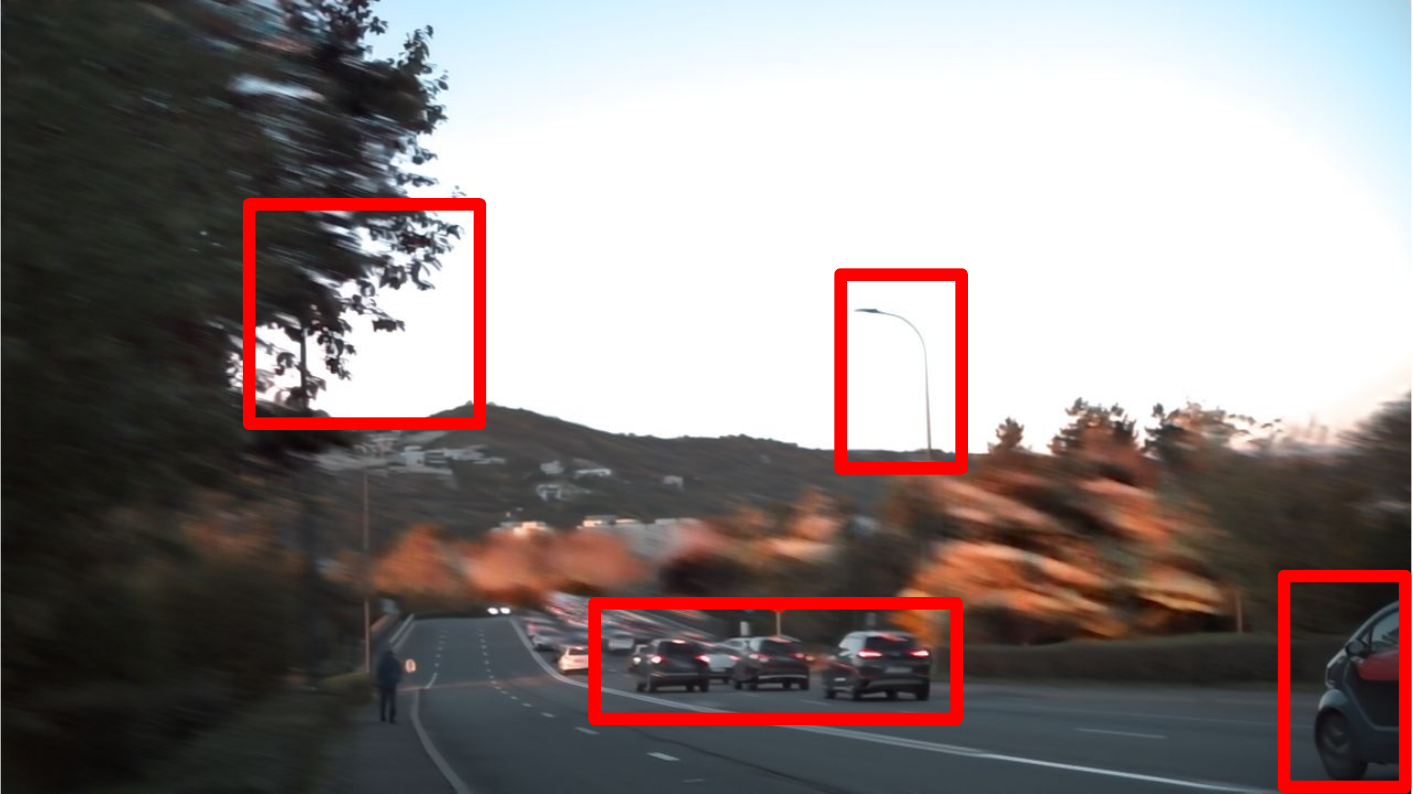}
    \includegraphics[width=.445\linewidth]{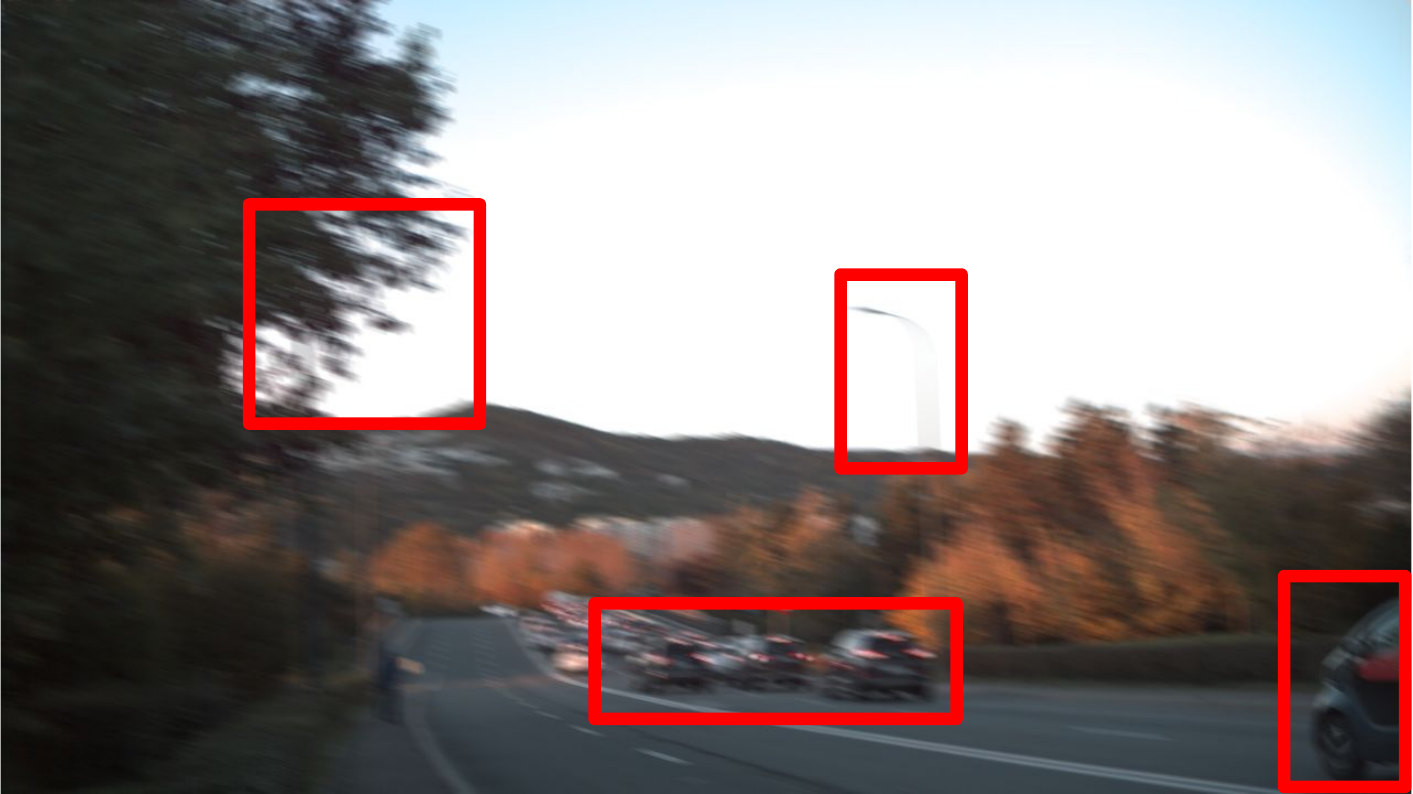}
    \caption{Fast Scene}
    \label{fig:1a}
  \end{subfigure}%
  \vspace*{\fill}
  \begin{subfigure}{.92\linewidth}
  \centering
    \includegraphics[width=.445\linewidth]{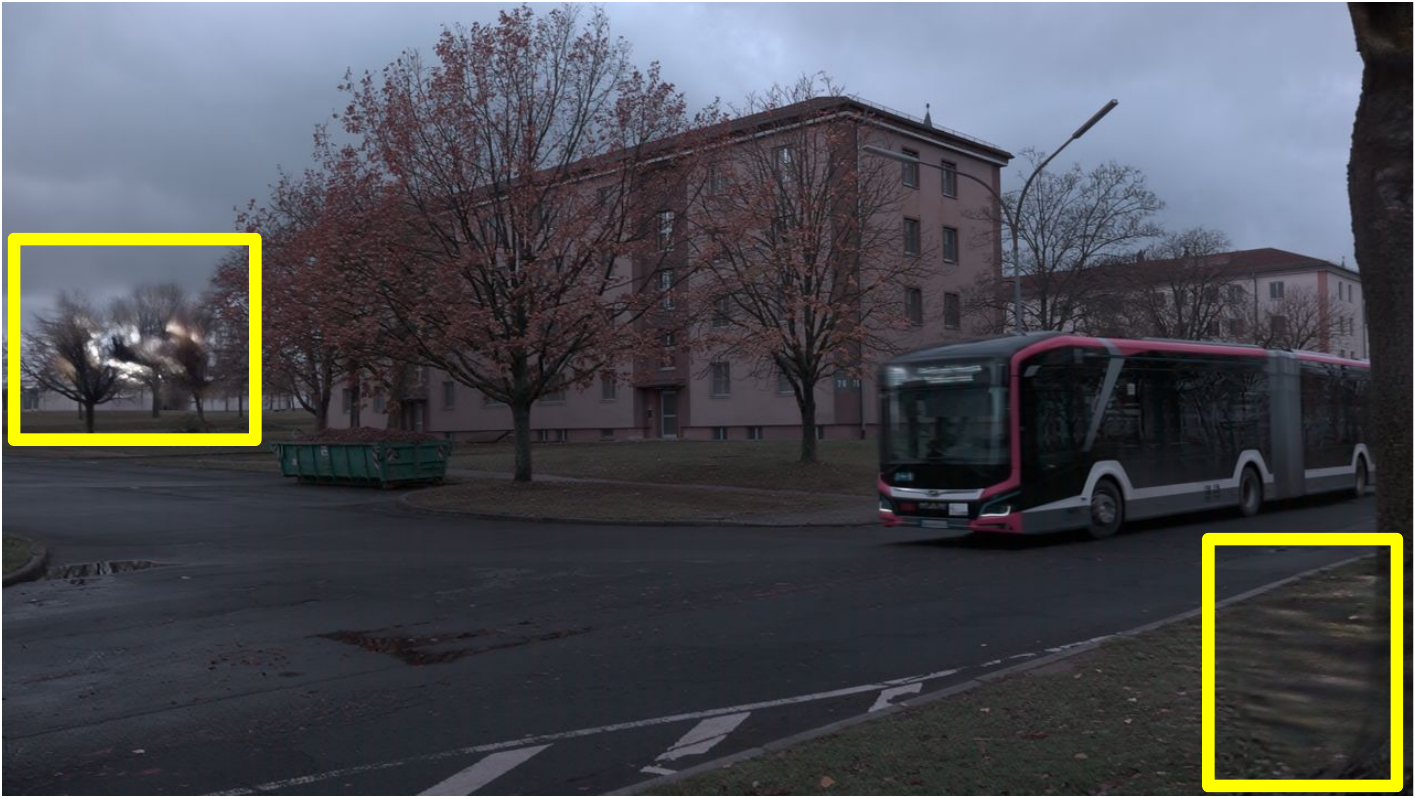}
    \includegraphics[width=.445\linewidth]
    {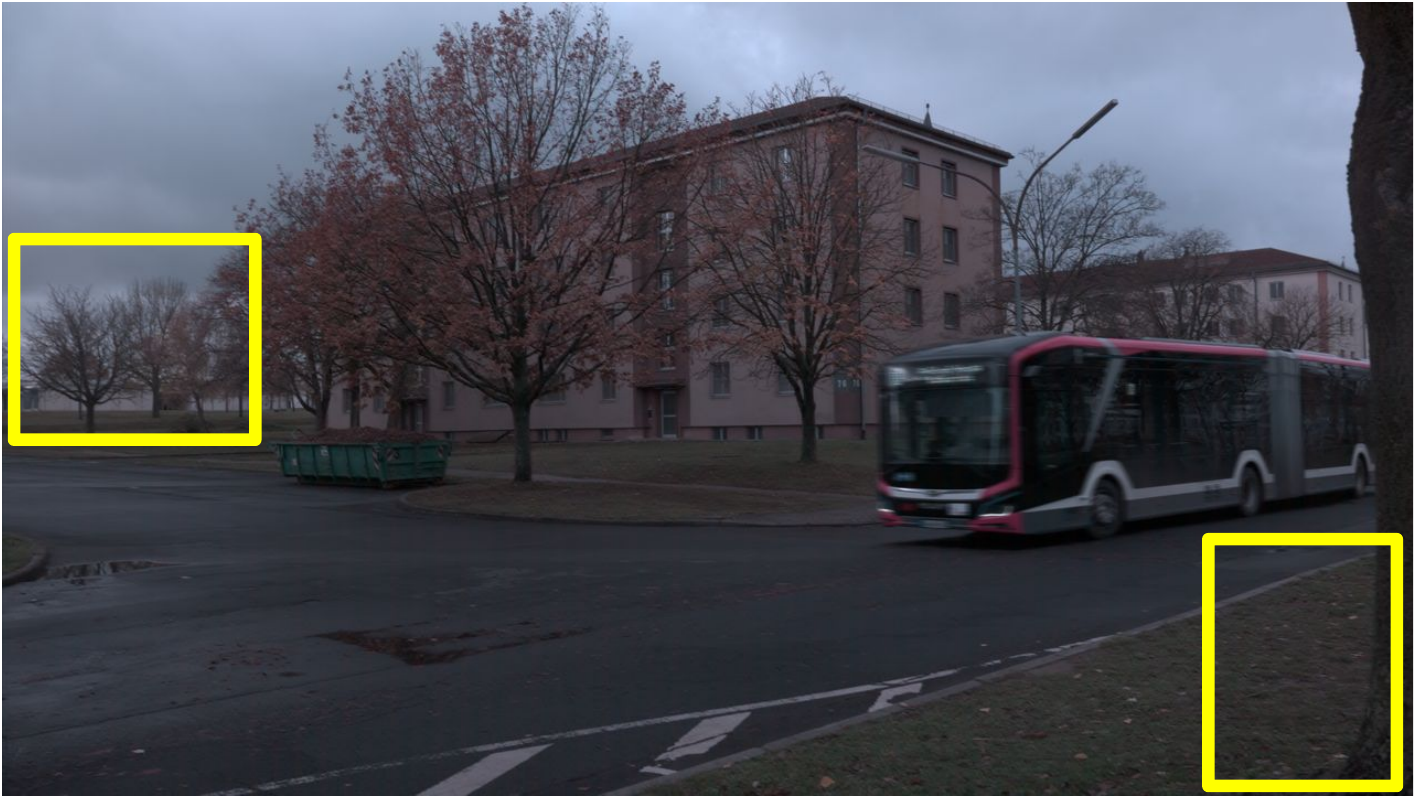}
    \caption{Slow Scene}
    \label{fig:1b}
  \end{subfigure}%
\caption{Motion Deblurring predictions of FFTformer (left) vs NAFNet (right). Although FFTformer better leverages contrast cues (as displayed within the \colorbox{red!30}{red} rectangles) to restore the high-intensity motion blur, NAFNet is more stable in slow scenes. The \colorbox{yellow!30}{yellow} rectangles focus on the sharp, high frequency features that are being misinterpreted by FFTformer. 
} \label{fig:out_FFTvsNAF}
\end{figure}

Additionally, as shown in Figure \ref{fig:out_FFTvsNAF}, both FFTformer and NAFNet excel in generalization across datasets. However, FFTformer outperforms NAFNet in fast motion but struggles with slow motion. This discrepancy is primarily due to frequency training, which enhances performance in fast motion but overcomplicates in simpler settings.

\textbf{Takeaways}: Fast motions are generally considered more challenging to handle. Priors such as frequency-based cues are often thought to be beneficial in these scenarios. While this is partially true, such methods tend to generalize poorly to slow-motion cases, often introducing hallucinations due to the mismatch between the prior and the actual blur characteristics. We venture that this highlights the need for a more principled mathematical modeling approach.

\begin{table}[t]
\resizebox{\linewidth}{!}{%
\begin{tabular}{c|ccccccc}
\textbf{Model} & XF & F & MF & M & MS & S & Total \\
\hline
AdaRevD & \cellcolor[HTML]{DAE8FC}31.79                  & \cellcolor[HTML]{DAE8FC}31.92                 & \cellcolor[HTML]{DAE8FC}32.22                  & \cellcolor[HTML]{DAE8FC}32.04                 & \cellcolor[HTML]{DAE8FC}30.09                  & \cellcolor[HTML]{DAE8FC}31.30                 & \cellcolor[HTML]{DAE8FC}31.70  \\
\cite{adarevd}               & 0.906                  & 0.899                 & 0.890                  & 0.901                 & 0.873                  & 0.893                 & 0.898 \\ \hline
LoFormer & \cellcolor[HTML]{DAE8FC}31.71                  & \cellcolor[HTML]{DAE8FC}31.77                 & \cellcolor[HTML]{DAE8FC}31.81                  & \cellcolor[HTML]{DAE8FC}32.20                 & \cellcolor[HTML]{DAE8FC}30.12                  & \cellcolor[HTML]{DAE8FC}31.32                 & \cellcolor[HTML]{DAE8FC}31.61 \\
\cite{loformer}               & 0.901                  & 0.894                 & 0.878                  & 0.904                 & 0.871                  & 0.890                 & 0.893 \\ \hline
FFTformer & \cellcolor[HTML]{DAE8FC}27.14                  & \cellcolor[HTML]{DAE8FC}27.80                 & \cellcolor[HTML]{DAE8FC}28.11                  & \cellcolor[HTML]{DAE8FC}28.52                 & \cellcolor[HTML]{DAE8FC}26.29                  & \cellcolor[HTML]{DAE8FC}26.89                 & \cellcolor[HTML]{DAE8FC}27.48 \\
\cite{fftformer}               & 0.853                  & 0.853                 & 0.832                  & 0.864                 & 0.812                  & 0.849                 & 0.848 \\ \hline
UFPNet & \cellcolor[HTML]{DAE8FC}31.48                  & \cellcolor[HTML]{DAE8FC}31.98                 & \cellcolor[HTML]{DAE8FC}32.66                  & \cellcolor[HTML]{DAE8FC}32.53                 & \cellcolor[HTML]{DAE8FC}30.62                  & \cellcolor[HTML]{DAE8FC}31.64                 & \cellcolor[HTML]{DAE8FC}31.78 \\
\cite{ufpnet}                & 0.900                    & 0.901                 & 0.896                  & 0.908                 & 0.883                  & 0.900                   & 0.899 \\ \hline
NAFNet & \cellcolor[HTML]{DAE8FC}26.73                  & \cellcolor[HTML]{DAE8FC}28.98                 & \cellcolor[HTML]{DAE8FC}31.12                  & \cellcolor[HTML]{DAE8FC}30.63                 & \cellcolor[HTML]{DAE8FC}29.36                  & \cellcolor[HTML]{DAE8FC}27.59                 & \cellcolor[HTML]{DAE8FC}28.56 \\
\cite{nafnet}               & 0.805                  & 0.851                 & 0.871                  & 0.887                 & 0.862                  & 0.837                 & 0.841

\end{tabular}%
}
\vspace{-1mm}
\captionof{table}{Reported \colorbox{tableblue!30}{PSNR} and SSIM of Single Image Motion Deblurring state-of-the-art methods on \emph{MIORe}. Their evaluation was performed on the following splits: \emph{XF} = Extreme, \emph{F} = Fast, \emph{MF} = Medium-Fast, \emph{M} = Medium, \emph{MS} = Mild-Medium, \emph{S} = Mild.}
    \label{table:results_deblurring}
    \vspace{-1mm}
\end{table}

\begin{table}[t]
\resizebox{\linewidth}{!}{%
\begin{tabular}{c|ccccccc}
\textbf{\# Frames}      & 1 & 5 & 13 & 29 & 61 & 125 & 249 \\ \hline
AdaRevD   & \cellcolor[HTML]{DAE8FC}39.61 & \cellcolor[HTML]{DAE8FC}34.03 & \cellcolor[HTML]{DAE8FC}31.39 & \cellcolor[HTML]{DAE8FC}28.42 & \cellcolor[HTML]{DAE8FC}25.83 & \cellcolor[HTML]{DAE8FC}23.27 & \cellcolor[HTML]{DAE8FC}21.39 \\
\cite{adarevd}& 0.990 & 0.933 & 0.893 & 0.834 & 0.781 & 0.721 & 0.672 \\ \hline
LoFormer  & \cellcolor[HTML]{DAE8FC}41.41 & \cellcolor[HTML]{DAE8FC}34.09 & \cellcolor[HTML]{DAE8FC}31.36 & \cellcolor[HTML]{DAE8FC}28.43 & \cellcolor[HTML]{DAE8FC}25.71 & \cellcolor[HTML]{DAE8FC}23.23 & \cellcolor[HTML]{DAE8FC}21.42 \\
\cite{loformer}& 0.988 & 0.931 & 0.890 & 0.835 & 0.779 & 0.724 & 0.679 \\ \hline
FFTformer & \cellcolor[HTML]{DAE8FC}35.33 & \cellcolor[HTML]{DAE8FC}29.17 & \cellcolor[HTML]{DAE8FC}27.34 & \cellcolor[HTML]{DAE8FC}25.72 & \cellcolor[HTML]{DAE8FC}24.20 & \cellcolor[HTML]{DAE8FC}22.41 & \cellcolor[HTML]{DAE8FC}20.91 \\
\cite{fftformer}& 0.972 & 0.891 & 0.848 & 0.807 & 0.768 & 0.723 & 0.684 \\ \hline
UFPNet    & \cellcolor[HTML]{DAE8FC}37.83 & \cellcolor[HTML]{DAE8FC}33.63 & \cellcolor[HTML]{DAE8FC}31.44 & \cellcolor[HTML]{DAE8FC}28.39 & \cellcolor[HTML]{DAE8FC}25.65 & \cellcolor[HTML]{DAE8FC}22.98 & \cellcolor[HTML]{DAE8FC}21.15 \\
\cite{ufpnet}& 0.983 & 0.931 & 0.891 & 0.837 & 0.785 & 0.727 & 0.684 \\ \hline
NAFNet    & \cellcolor[HTML]{DAE8FC}25.69 & \cellcolor[HTML]{DAE8FC}28.25 & \cellcolor[HTML]{DAE8FC}27.99 & \cellcolor[HTML]{DAE8FC}26.47 & \cellcolor[HTML]{DAE8FC}24.19 & \cellcolor[HTML]{DAE8FC}21.87 & \cellcolor[HTML]{DAE8FC}20.03 \\
\cite{nafnet}& 0.805 & 0.846 & 0.836 & 0.799 & 0.752 & 0.701 & 0.655 \\ \hline
baseline     & \cellcolor[HTML]{DAE8FC}inf   & \cellcolor[HTML]{DAE8FC}34.32 & \cellcolor[HTML]{DAE8FC}30.05 & \cellcolor[HTML]{DAE8FC}27.13 & \cellcolor[HTML]{DAE8FC}24.88 & \cellcolor[HTML]{DAE8FC}23.03 & \cellcolor[HTML]{DAE8FC}21.52 \\
input        & 1.000     & 0.920 & 0.863 & 0.815 & 0.775 & 0.738 & 0.704
\end{tabular}%
}
\vspace{-1mm}
\captionof{table}{{\colorbox{tableblue!30}{PSNR} and SSIM for the Motion Deblurring state-of-the-art methods on our \emph{VAR-MIORe} dataset. \emph{VAR-MIORe-Deblurring} is organized in buckets from 1 to 249, specifying the number of images averaged for creating the blur effect in this case. The data consists of 333 input entries per bucket. All the results have been computed and this table summarizes the overall trend. The head of the table tells us the number of frames averaged to achieve the variable intensity blurry frames, given as input to the models.}}
\label{table:psnr_varmiore_blur}
\vspace{-3mm}
\end{table}

\subsection{Video Frame Interpolation}

Next, we assess VFI performance. In this task, two sharp frames serve as inputs, with the intervening middle frame used as ground truth. We select the best available checkpoints for each of the five VFI methods. Performance on \emph{MIORe} is summarized in Table \ref{table:results_vfi}, while results on \emph{VAR-MIORe} are reported in Table \ref{table:results_var_vfi} and visualized in Figures from the \emph{supplementary material}. Surprisingly, our observations indicate that all methods perform well under extreme motion conditions; however, their performance deteriorates when interpolating moderate motion scenarios.

\begin{table}[t]
\footnotesize
\resizebox{\linewidth}{!}{%
\begin{tabular}{c|ccccccc}
\textbf{Model} & XF & F & MF & M & MS & S & Total \\
\hline
VFIMamba       & \cellcolor[HTML]{DAE8FC}35.67                  & \cellcolor[HTML]{DAE8FC}34.55                 & \cellcolor[HTML]{DAE8FC}34.53                  & \cellcolor[HTML]{DAE8FC}34.32                 & \cellcolor[HTML]{DAE8FC}31.86                  & \cellcolor[HTML]{DAE8FC}30.29                 & \cellcolor[HTML]{DAE8FC}34.41 \\
\cite{vfimamba}               & 0.940                   & 0.930                  & 0.929                  & 0.931                 & 0.911                  & 0.882                 & 0.930  \\ \hline
SGM-VFI        & \cellcolor[HTML]{DAE8FC}35.60                   & \cellcolor[HTML]{DAE8FC}34.14                 & \cellcolor[HTML]{DAE8FC}34.29                  & \cellcolor[HTML]{DAE8FC}34.25                 & \cellcolor[HTML]{DAE8FC}31.81                  & \cellcolor[HTML]{DAE8FC}30.43                 & \cellcolor[HTML]{DAE8FC}34.24 \\
\cite{sgmvfi}               & 0.940                   & 0.926                 & 0.927                  & 0.930                  & 0.911                  & 0.885                 & 0.928 \\ \hline
PerVFI         & \cellcolor[HTML]{DAE8FC}31.69                  & \cellcolor[HTML]{DAE8FC}30.77                 & \cellcolor[HTML]{DAE8FC}30.86                  & \cellcolor[HTML]{DAE8FC}31.17                 & \cellcolor[HTML]{DAE8FC}29.43                  & \cellcolor[HTML]{DAE8FC}28.31                 & \cellcolor[HTML]{DAE8FC}30.88 \\
\cite{pervfi}               & 0.877                  & 0.866                 & 0.858                  & 0.873                 & 0.852                  & 0.827                 & 0.866 \\ \hline
EMA-VFI        & \cellcolor[HTML]{DAE8FC}35.60                   & \cellcolor[HTML]{DAE8FC}34.30                  & \cellcolor[HTML]{DAE8FC}34.43                  & \cellcolor[HTML]{DAE8FC}34.26                 & \cellcolor[HTML]{DAE8FC}32.09                  & \cellcolor[HTML]{DAE8FC}30.68                 & \cellcolor[HTML]{DAE8FC}34.35 \\
\cite{emavfi}               & 0.938                  & 0.926                 & 0.931                  & 0.930                  & 0.913                  & 0.888                 & 0.928 \\ \hline
BiFormer       & \cellcolor[HTML]{DAE8FC}33.47                  & \cellcolor[HTML]{DAE8FC}32.57                 & \cellcolor[HTML]{DAE8FC}32.73                  & \cellcolor[HTML]{DAE8FC}32.79                 & \cellcolor[HTML]{DAE8FC}31.05                  & \cellcolor[HTML]{DAE8FC}29.42                 & \cellcolor[HTML]{DAE8FC}32.60  \\
\cite{biformer}               & 0.915                  & 0.906                 & 0.904                  & 0.909                 & 0.895                  & 0.866                 & 0.906
\end{tabular}%
}
\vspace{-1mm}
\captionof{table}{\colorbox{tableblue!30}{PSNR} and SSIM performances of Video Frame Interpolation state-of-the-art methods on our dataset, \emph{MIORe}. 
}
    \label{table:results_vfi}
    \vspace{-3mm}
\end{table}

\begin{table}[t]
\resizebox{\linewidth}{!}{%
\begin{tabular}{c|ccccccc}
\textbf{\# Frames}      & 1 & 5 & 13 & 29 & 61 & 125 & 249 \\ \hline
VFIMamba          & \cellcolor[HTML]{DAE8FC}67.79 & \cellcolor[HTML]{DAE8FC}37.18 & \cellcolor[HTML]{DAE8FC}34.24 & \cellcolor[HTML]{DAE8FC}28.68 & \cellcolor[HTML]{DAE8FC}23.56 & \cellcolor[HTML]{DAE8FC}20.58 & \cellcolor[HTML]{DAE8FC}18.91 \\
\cite{vfimamba}                  & 0.999                         & 0.950                         & 0.928                         & 0.839                         & 0.730                         & 0.663                         & 0.621                         \\ \hline
SGM-VFI           & \cellcolor[HTML]{DAE8FC}56.80 & \cellcolor[HTML]{DAE8FC}37.06 & \cellcolor[HTML]{DAE8FC}33.70 & \cellcolor[HTML]{DAE8FC}28.69 & \cellcolor[HTML]{DAE8FC}23.94 & \cellcolor[HTML]{DAE8FC}20.68 & \cellcolor[HTML]{DAE8FC}18.71 \\
\cite{sgmvfi}                  & 0.999                         & 0.951                         & 0.921                         & 0.839                         & 0.732                         & 0.652                         & 0.598                         \\ \hline
PerVFI            & \cellcolor[HTML]{DAE8FC}46.59 & \cellcolor[HTML]{DAE8FC}32.75 & \cellcolor[HTML]{DAE8FC}30.80 & \cellcolor[HTML]{DAE8FC}27.60 & \cellcolor[HTML]{DAE8FC}23.78 & \cellcolor[HTML]{DAE8FC}20.74 & \cellcolor[HTML]{DAE8FC}18.63 \\
\cite{sgmvfi}                  & 0.996                         & 0.891                         & 0.869                         & 0.808                         & 0.717                         & 0.645                         & 0.592                         \\ \hline
EMA-VFI           & \cellcolor[HTML]{DAE8FC}81.16 & \cellcolor[HTML]{DAE8FC}37.06 & \cellcolor[HTML]{DAE8FC}34.08 & \cellcolor[HTML]{DAE8FC}29.07 & \cellcolor[HTML]{DAE8FC}24.00 & \cellcolor[HTML]{DAE8FC}20.78 & \cellcolor[HTML]{DAE8FC}18.91 \\
\cite{emavfi}                  & 0.999                         & 0.949                         & 0.926                         & 0.851                         & 0.745                         & 0.676                         & 0.630                         \\ \hline
BiFormer          & \cellcolor[HTML]{DAE8FC}43.60 & \cellcolor[HTML]{DAE8FC}34.57 & \cellcolor[HTML]{DAE8FC}32.38 & \cellcolor[HTML]{DAE8FC}28.79 & \cellcolor[HTML]{DAE8FC}24.36 & \cellcolor[HTML]{DAE8FC}20.94 & \cellcolor[HTML]{DAE8FC}18.90 \\
\cite{biformer}                  & 0.992                         & 0.928                         & 0.906                         & 0.843                         & 0.746                         & 0.672                         & 0.620                         \\ \hline
baseline & \cellcolor[HTML]{DAE8FC}inf   & \cellcolor[HTML]{DAE8FC}30.41 & \cellcolor[HTML]{DAE8FC}25.92 & \cellcolor[HTML]{DAE8FC}23.14 & \cellcolor[HTML]{DAE8FC}21.09 & \cellcolor[HTML]{DAE8FC}19.45 & \cellcolor[HTML]{DAE8FC}18.22 \\
AVG(L, R)                  & 1.000                             & 0.839                         & 0.753                         & 0.694                         & 0.646                         & 0.604                         & 0.566                        
\end{tabular}%
}
\vspace{-1mm}
\captionof{table}{\colorbox{tableblue!30}{PSNR} and SSIM for the Video Frame Interpolation state-of-the-art methods on \emph{VAR-MIORe}. Table head mentions the displacement between the left and right frames given as input to the models. The baseline method consists of comparing the average of the two input images to the middle ground truth frame.}
\label{table:results_var_vfi}
\end{table}

\subsection{Optical Flow Estimation}
\label{ssec:of-discussion}

Lastly, we present preliminary results for optical flow estimation. Table \ref{table:results_of} compares the performance of five methods evaluated on pseudo-ground truth labels generated following the approach in \cite{deqflow}. Although these results serve as a proof of concept to demonstrate the potential of our dataset, further refinement of the ground truth is required. This initial evaluation aims to stimulate further research into developing an omni-task dataset for motion analysis. Further details regarding the optical flow estimation algorithm within the frame of blur generation, and the limitations of our incipient work are presented in the \emph{supplementary material}.

\begin{table}[t]
\footnotesize
\resizebox{\linewidth}{!}{%
\begin{tabular}{c|ccccccc}
\textbf{Model} & XF                            & F                             & MF                           & M                            & MS                           & S                            & Total                         \\ \hline
VideoFlow      & \cellcolor[HTML]{DAE8FC}6.17  & \cellcolor[HTML]{DAE8FC}8.44  & \cellcolor[HTML]{DAE8FC}9.12 & \cellcolor[HTML]{DAE8FC}7.69 & \cellcolor[HTML]{DAE8FC}9.57 & \cellcolor[HTML]{DAE8FC}8.88 & \cellcolor[HTML]{DAE8FC}19.38 \\
\cite{videoflow}               & 52.43                         & 67.64                         & 75.64                        & 69.12                        & 76.93                        & 59.29                        & 63.74                         \\ \hline
FlowFormer     & \cellcolor[HTML]{DAE8FC}33.81 & \cellcolor[HTML]{DAE8FC}4.06  & \cellcolor[HTML]{DAE8FC}3.94 & \cellcolor[HTML]{DAE8FC}3.42 & \cellcolor[HTML]{DAE8FC}4.29 & \cellcolor[HTML]{DAE8FC}4.41 & \cellcolor[HTML]{DAE8FC}14.65 \\
\cite{flowformer}               & 27.64                         & 38.71                         & 44.15                        & 41.93                        & 48.45                        & 33.81                        & 36.45                         \\ \hline
FlowDiffuser   & \cellcolor[HTML]{DAE8FC}4.55  & \cellcolor[HTML]{DAE8FC}6.61  & \cellcolor[HTML]{DAE8FC}8.21 & \cellcolor[HTML]{DAE8FC}7.15 & \cellcolor[HTML]{DAE8FC}6.09 & \cellcolor[HTML]{DAE8FC}5.97 & \cellcolor[HTML]{DAE8FC}6.01  \\
\cite{flowdiffuser}               & 38.53                         & 52.34                         & 59.22                        & 56.51                        & 61.92                        & 45.71                        & 49.25                         \\ \hline
MemFlow        & \cellcolor[HTML]{DAE8FC}2.96  & \cellcolor[HTML]{DAE8FC}4.35  & \cellcolor[HTML]{DAE8FC}2.66 & \cellcolor[HTML]{DAE8FC}3.08 & \cellcolor[HTML]{DAE8FC}0.67 & \cellcolor[HTML]{DAE8FC}1.50 & \cellcolor[HTML]{DAE8FC}2.99  \\
\cite{memflow}               & 4.49                          & 9.13                          & 7.14                         & 6.51                         & 2.95                         & 3.50                         & 6.01                          \\ \hline
Ef-RAFT        & \cellcolor[HTML]{DAE8FC}4.68  & \cellcolor[HTML]{DAE8FC}10.64 & \cellcolor[HTML]{DAE8FC}8.03 & \cellcolor[HTML]{DAE8FC}7.02 & \cellcolor[HTML]{DAE8FC}8.15 & \cellcolor[HTML]{DAE8FC}9.88 & \cellcolor[HTML]{DAE8FC}7.52  \\
\cite{efraft}               & 12.44                         & 20.22                         & 17.73                        & 16.66                        & 16.86                        & 15.83                        & 16.18                        
\end{tabular}%
}
\vspace{-1mm}
\captionof{table}{Reported \colorbox{tableblue!30}{EPE} and F1 performances of Optical Flow Estimation state-of-the-art methods on our dataset, \emph{MIORe}. 
}
\label{table:results_of}
\vspace{-3mm}
\end{table}

\section{Conclusions}
\label{sec:conclusions}

In this paper, we have presented \emph{MIORe} and \emph{VAR-MIORe}, two novel multi-task datasets that capture a broad spectrum of motion scenarios and intensities. \emph{MIORe} provides a robust and fair multi-task benchmark with flow-based regularized motion blur that mirrors real-world imaging conditions.
In contrast, \emph{VAR-MIORe} extends this capability by incorporating a variable range of motion magnitudes, from minimal to extreme, thereby establishing the first dataset of its kind for exploring the limits of motion deblurring, video interpolation, and optical flow estimation. Our extensive benchmark evaluations reveal the performance boundaries of state-of-the-art approaches under these diverse and challenging conditions. Together, these datasets not only address critical gaps in the current literature but also open new avenues for the development of advanced, multi-task restoration algorithms capable of handling a wide array of optical degradations.

\noindent \textbf{Acknowledgements.}
The authors sincerely thank the reviewers and all members of the program committee for their tremendous efforts and incisive feedback. 
This research was supported by the Alexander von Humboldt Foundation.

{
    \small
    \bibliographystyle{ieeenat_fullname}
    \bibliography{main}
}

\end{document}